\documentclass[11pt,a4paper]{article}
\usepackage[hyperref]{naaclhlt2018}
\usepackage{times}
\usepackage{latexsym}
\usepackage{latexsym}
\usepackage{amssymb}
\usepackage{amstext}
\usepackage{amsmath}
\usepackage{graphicx}
\usepackage{url}
\usepackage{multirow}

\aclfinalcopy % Uncomment this line for all SemEval submissions

\title{EiTAKA at SemEval-2018 Task 1: An Ensemble of N-Channels ConvNet and XGboost Regressors for Emotion Analysis of Tweets}
\author{Mohammed Jabreel\qquad   Antonio Moreno \\ Intelligent Technologies for Advanced Knowledge Acquisition (ITAKA), \\ Departament d'Enginyeria Inform\`{a}tica i Matem\`{a}tiques, \\ Universitat Rovira i Virgili,\\	Av. Pa\"{i}sos Catalans, 26, 43007 Tarragona, Spain \\
	\textit{$<$first\_name$>$.$<$last\_name$>$@urv.cat}}

\date{}

\begin{document}
\maketitle
\begin{abstract}

This paper describes our system that has been used in Task1 Affect in Tweets. We combine two different approaches. The first one called N-Stream ConvNets, which is a deep learning approach where the second one is XGboost regresseor based on set of embedding and lexicons based features. Our system was evaluated on the testing sets of the tasks outperforming all other approaches for the Arabic version of valence intensity regression task and valence ordinal classification task. 
  
\end{abstract}

\section{Introduction}

\textit{Sentiment Analysis} is the task of automatically identifying the valence or polarity of a piece of text. This piece of text can be a user review, a document, an SMS message, a tweet, etc. According to \cite{SentimentEmotionSurvey2015}, the term sentiment analysis also refers to determining the attitude towards a particular target or topic. The attitude can be the polarity (positive or negative), or an emotional or effectual attitude such as joy, anger, sadness and so on.

Most of the researchers in sentiment analysis have focused on developing systems to determine the polarity of a given text. This involves designing classifiers based on a set of examples with a manually annotated sentiment polarity. Although developing systems that automatically determine the intensity (i.e. the degree or the amount) of emotions that are communicated in a text has a wide range of applications in commerce, public health, social welfare, etc.,  most of the work has focused on categorical classification (whether a given piece of text communicates anger, joy, sadness, etc.). This can be attributed to the lack of suitable annotated data \cite{wassa17} .

In task1: \textit{Affect in Tweets}, the organizers provide an array of tasks where systems have to automatically determine the intensity of emotions (anger, fear, joy, and sadness) and the intensity of the sentiment (aka valence) of the tweeters from their tweets. They provide annotated datasets for each task with English, Arabic, and Spanish tweets \cite{SemEval2018Task1}. We define the tasks below:

\textbf{EI-reg} (an emotion intensity regression task): Given a tweet and an emotion $E$, determine the  intensity of $E$ that best represents the mental state of the tweeter with a real-valued score between 0 (least $E$) and 1 (most $E$).

\textbf{EI-oc} (an emotion intensity ordinal classification task): Given a tweet and an emotion $E$, classify the tweet into one of four ordinal classes of intensity of $E$ that best represents the mental state of the tweeter.

\textbf{V-reg} (a sentiment intensity regression task): Given a tweet, determine the intensity of sentiment or valence $V$ that best represents the mental state of the tweeter with a real-valued score between 0 (most negative) and 1 (most positive).

\textbf{V-oc} (a sentiment analysis, ordinal classification, task): Given a tweet, classify it into one of seven ordinal classes, corresponding to various levels of positive and negative sentiment intensity, that best represents the mental state of the tweeter.

We proposed one system to solve the intensity regression tasks (i.e. EI-reg and V-reg) and use it as a feature extractor to train Decision Trees to solve the ordinal classification tasks (i.e. EI-oc and V-oc). We developed two versions of the proposed system for the English and the Arabic language tweets.
 
Our system is an ensemble of two different approaches. The first one, called N-Channels ConvNet, is a deep learning approach where the second one is an XGboost regressor based on a set of embedding and lexicons-based features.

The rest of the paper is organized as follows: Section 2 presents the tools and the resources that are used. Section 3 describes the proposed system. In Section 4 we report the experimental results, whereas in Section 5 the conclusions and the future work are presented.

\section{Resources}

This section explains the tools and the resources that have been used in our system. 

\subsection{Sentiment Lexicons}
We used the following lexicons for the English version of our system:

AFINN \cite{afinn}, General Inquirer \cite{stone1968general}, Bing-Liu opinion lexicon (HL) \cite{hu2004mining}, MPQA \cite{choi2014+}, NRC hashtag sentiment lexicon \cite{MohammadKZ2013}, NRC emotion lexicon (EmoLex), NRC affect intensity lexicon, NRC hashtag emotion lexicon and Vader lexicon. More details about each lexicon, such as how it was created, the polarity score for each term, and the statistical distribution of the lexicon, can be found in \cite{JabreelM16}.

For the Arabic version we used the following lexicons:

Arabic Hashtag lexicon, Dialectal Arabic Hashtag lexicon, Arabic Bing Liu lexicon, Arabic Sentiment140 lexicon and Arabic translation of the NRC emotion lexicon. The first two were created manually, whereas the rest were translated to Arabic from the English version using Google Translator. \cite{arabiclex2016}.

\subsection{Embeddings}

{\sl Word embeddings} are an approach for distributional semantics which represents words as vectors of real numbers. Such representation has useful clustering properties, since the words that are semantically and syntactically related are represented by similar vectors \cite{mikolov2013efficient}. For example, the words "coffee" and "tea" will be very close in the created space.

We used two publicly available pre-trained embedding models in the English version of our system. The first one was used in \cite{rouvier2016sensei}. It was trained using word2vec (skipgram model) on an unannotated corpus of 20 million English tweets containing at least one emoticon.
%We call it SENSEI-LIF.
The second one was provided by \cite{baziotis17}.
%and we call it Glove-DS.
It was trained on a big dataset of 330M English Twitter messages, gathered from 12/2012 to 07/2016 and a vocabulary size of 660K words using Glove algorithm.

Additionally, we have trained two embedding models on 60M English tweets(30M contain positive emoticons, 30M negative ones). The first one was trained by applying word2vec skipgram of window size 5 and filtering words that occur less than 4 times. The dimensionality of the vector was set to 300. The second one was trained using fastText [CITE]. The dimensionality of the vector was set to 300.

Similarly, we used two publicly available pre-trained embedding models in the Arabic version of our system and trained two. The first one is the model Arabic-SKIP-G300, provided by \cite{zahran2015word}. Arabic-SKIP-G300 was trained on a large corpus of Arabic text collected from different sources such as Arabic Wikipedia, Arabic Gigaword Corpus, Ksucorpus, King Saud University Corpus, Microsoft crawled Arabic Corpus, etc. It contains 300-dimensional vectors for 6M words and phrases. The second one is Twitter-SG-AraVec \cite{SOLIMAN2017256}, which was trained using word2vec skipgram algorithm on 66M Arabic tweets and 1B tokens. The dimensionality of the vector was set to 300. 

Our embedding models were trained on the distant supervision corpus (about 16M Arabic tweets) provided by the organizers. We were able to find  about 12M tweets. Again, similar to our English embeddings, we trained the two Arabic embedding models.

\section{System Description}
This section explains the proposed system, whose architecture is shown in Figure 1. First, we preprocess the tweets (Subsection 3.1). Afterwards, we pass them to the N-Channels ConvNet and the XGboost regressors (Subsections 3.2 and 3.3). Finally we ensemble the output of the two systems to get the final result as described in subsection 3.4. The proposed system is also used as feature extractor to train an ordinal Decision Tree classifier. as described in subsection 3.5.

\begin{figure}[h]
	\centering
	\includegraphics[width=0.5\textwidth]{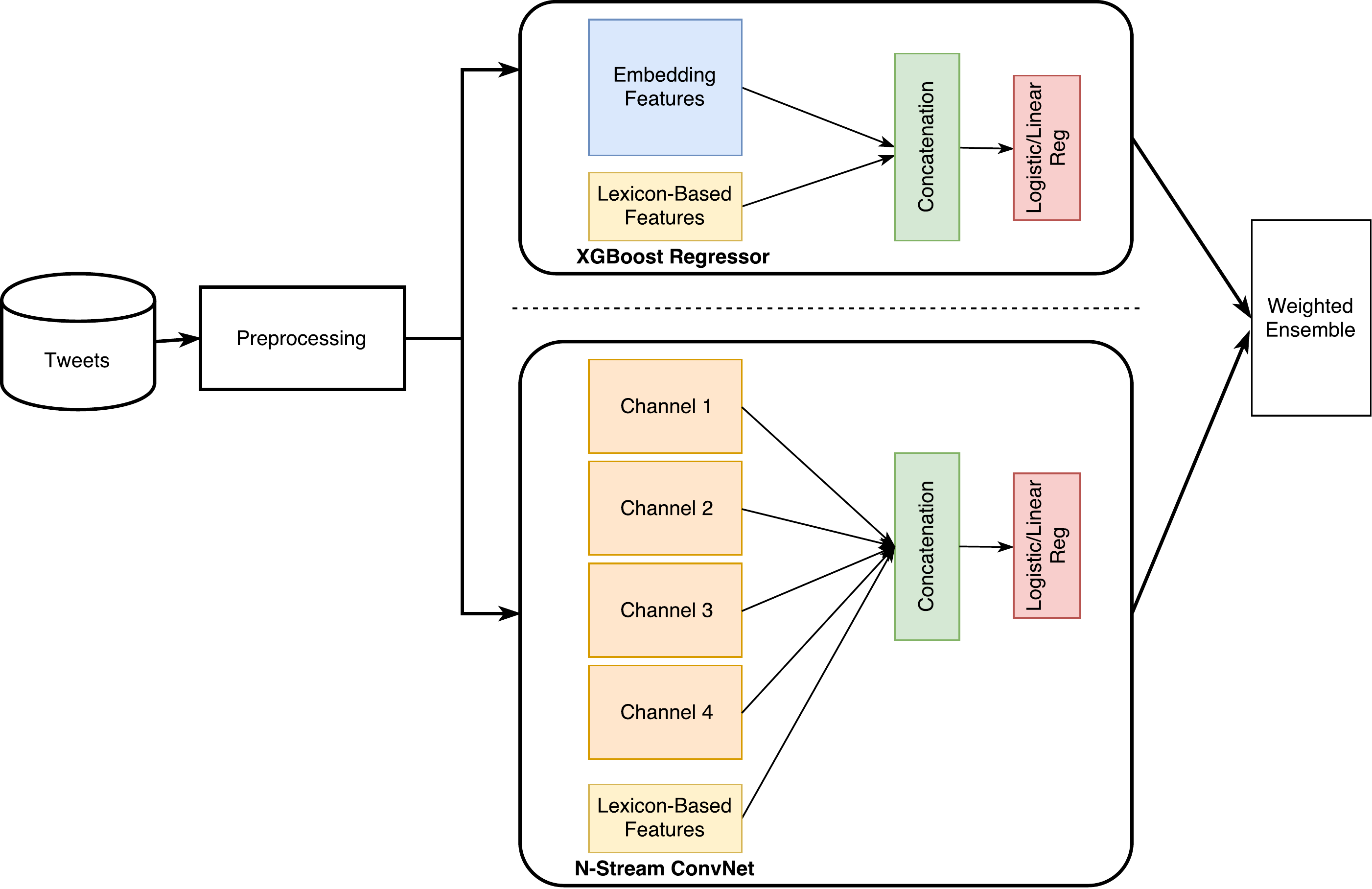}
	\caption{System Architecture.}
\end{figure}

\subsection{Preprocessing}

Some standard pre-processing methods were applied on the tweets:
\begin{itemize}
	\item \textit{Normalization}: Each tweet in English was converted to the lowercase.
	URLs and usernames were omitted. Non-Arabic letters were removed from each tweet in the Arabic-language sets. Words with repeated letters (i.e. elongated) are corrected. 
	
	\item \textit{Tokenization and POS tagging}: All English-language tweets were tokenized and tagged using Ark Tweet NLP \cite{Gimpel2011}, while all Arabic-language tweets were tokenized and tagged using Stanford Tagger \cite{green2010better}.
\end{itemize}

\subsection{N-Channels ConvNet}
Convolutional Neural Networks (ConvNets) have achieved remarkable results in computer vision and speech recognition tasks in recent years. The next subsection explains the architecture of our proposed ConvNet.
\subsubsection{Architecture}
 The N-Channels ConvNet model architecture, shown in the bottom box in figure 1, is inspired by Inception-Net \cite{inception4} and the CNN proposed by \cite{kim2014convolutional}. It is composed of multiple channels followed by a logistic regressor. Figure 2 shows the channel architecture.
The input to each channel is a sequence of words $w_1, w_2, ... w_n$ where $n$ is the number of words. We pass the input through an embedding layer to map each word $w_i$ into a real-valued vector. Each channel has its own embedding layer which is initialized by a specific pre-trained embedding model. We use five channels with the four pre-trained embedding models described in subsection 2.2 and a character based one. 
%Each channel in the joy emotion model is followed by a LeakyReLU pre-activation layer. The channels of the fear emotion model are followed by a projection dense layer with ReLU activation. 
The result from the embedding layer is a matrix $n \times d_c$ where $d_c$ is the vector dimension. This matrix is passed to a projection or pre-activation layer. Afterward, we feed the projected matrix to three Conv1D. Each one has a different kernel (1, 2, and 3) and 200 filters. To get more details about the architecture of this Conv1D please check \cite{kim2014convolutional}. We pass the output of each Conv1D through a global max-pooling layer which produces a vector with dimensionality of 200. Finally, the three vectors are concatenated. This yields a vector with dimensionality of 600 that represents the tweet (i.e. the input sequence of words).

\begin{figure}[h]
	\centering
	\includegraphics[width=0.4\textwidth, height=0.6\textheight]{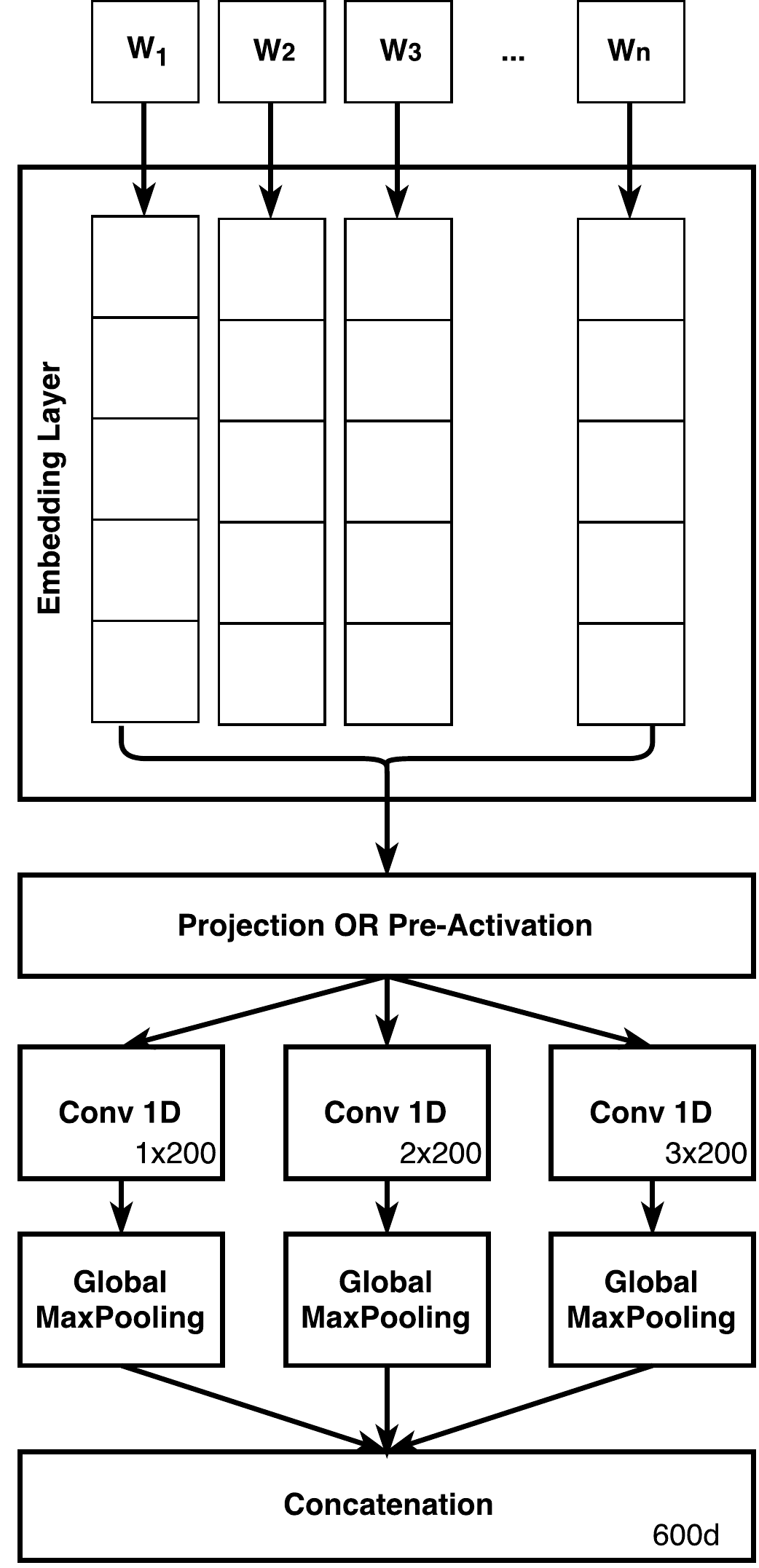}
	\caption{Channel Architecture.}
\end{figure}

Finally, the outputs of all channels are concatenated with a lexicon-based vector (see next section) and fed to a single sigmoid neuron which gives the intensity of the emotion/valence.

\subsubsection{Training}
The proposed model was trained by minimizing the mean squared error between the real and predicted intensities. The optimization was done by applying back-propagation through layers via minibatch gradient descent. The training parameters were the following: batch size of 32, 100 epochs and Adam optimization method with learning rate of 0.001, $\beta_1$ = 0.9 and $\beta_2$ = 0.999 and $\epsilon$ = $10^-9$. To prevent the over-fitting, we used dropout and early stopping methods. 

\subsection{XGBoost Regressor}

XGBoost \cite{xgb} has become a widely used and really popular tool among Data Scientists in industry, as it shows great performance on large-scale problems. It is a highly flexible and versatile tool that can work through most regression, classification and ranking problems as well as user-built objective functions.

We trained an XGBoost regressor to give the intensity of the emotion/valence based on the two types of features explained in the next subsection. 

\subsubsection{Features} Each tweet is represented with a vector by concatenating the following two feature vectors:

\textbf{Lexicon Features}: For each lexicon, we used the sum of the scores provided by the lexicon for each word in the tweet. Let $L$ denote the set of lexicons and $f_i^l(w)$ the score of the word $w$ based on the feature $i$ in the lexicon $l$ (note that some lexicons have only one feature like the sentiment score and some of them have multiple features like anger emotion score, positive score, etc). Then, the set of features that represent a given tweet $T$  and a lexicon $l \in L$ can be obtained as follows:

\begin{equation}
	V_{T, l} = \forall_{f_i^l \in F_l} \sum_{w \in T} f_i^l(w)
\end{equation}

Here, $F_l$ denotes the set of features in lexicon $l$.

\textbf{Embedding Features}: 
We used the \textit{sum} pooling function to obtain the tweet representation in the embedding space. More formally, let us consider an embedding matrix $E \in \mathbb{R}^{d \times |V|}$ and a tweet $T = {w_1, w_2, ..., w_n}$, where $d$ is the dimension size, $|V|$ is the length of the vocabulary (i.e. the number of words in the embedding model), $w_i$ is $i$-th the word in the tweet and $n$ is the number of words. First, each word $w_i$ is substituted by the corresponding vector $v_i^j$ in the matrix $E$ where $j$ is the index of the word $w_i$ in the vocabulary. This step ends with the matrix $W \in \mathbb{R}^{d \times n}$. The vector $V_{T, E}$ that represents the tweet $T$ is computed by aggregating the matrix $W$. This aggregation is done by taking the summation over its columns.
The sum spooling function is an element-wise function, and it converts texts with various lengths into a fixed-length vector allowing to capture the information throughout the entire text. 

\subsubsection{Training}
The XGBoost regressor has some parameters that need to be tuned. Table 1 shows the values of each parameter we chose for the different emotions. All those values were chosen using the grid-search on the development sets.

\begin{table}[h]
	\centering
	
	\label{tblxgb}
	\begin{tabular}{|l|l|l|l|l|l|}
		\hline
		& P       & \# Est. & S    & M & O        \\ \hline
		\multirow{5}{*}{Eng.} & Anger   & 300     & 0.75 & 5 & Logistic \\ \cline{2-6} 
		& Fear    & 300     & 0.75 & 5 & Linear   \\ \cline{2-6} 
		& Sadness & 300     & 0.75 & 5 & Logistic \\ \cline{2-6} 
		& Joy     & 300     & 0.75 & 7 & Linear   \\ \cline{2-6} 
		& Valence & 300     & 0.75 & 5 & Linear   \\ \hline \hline
		\multirow{5}{*}{Ara.}  & Anger   & 200     & 0.9  & 9 & Logistic \\ \cline{2-6} 
		& Fear    & 200     & 0.9  & 5 & Logistic \\ \cline{2-6} 
		& Sadness & 200     & 0.9  & 5 & Logistic \\ \cline{2-6} 
		& Joy     & 200     & 0.9  & 5 & Logistic \\ \cline{2-6} 
		& Valence & 200     & 0.9  & 9 & Logistic \\ \hline
	\end{tabular}
	\caption{The XGBoost regressors parameters. \#Est. refers to the number of estimators, S is the subsample, M is the maximum depth and O refers to the objective function.}
\end{table}

\subsection{Ensemble}
We combined the results of the two systems described above with the intention of improving the performance and increasing the generalizability of the final system. We used the weighted average method to achieve that. Let $r_1$ and $r_2$ respectively denote the output of the XGBoost regressor and the N-Channels ConvNet system. The final output $r$ was obtained as follows:
\begin{equation}
	r = \alpha * r_1 + (1 - \alpha) * r_2; \;\;\; \alpha \in [0, 1]
\end{equation} 

% Please add the following required packages to your document preamble:
% \usepackage{multirow}
\begin{table}[]
	\centering
	
	\label{alpha-vals}
	\begin{tabular}{|l|l|l|}
		\hline
		& Emotion & $\alpha$ \\ \hline
		\multirow{5}{*}{English} & Anger   & 0.3    \\ \cline{2-3} 
		& Fear    & 0.5    \\ \cline{2-3} 
		& Sadness & 0.6    \\ \cline{2-3} 
		& Joy     & 0.2    \\ \cline{2-3} 
		& Valence & 0.6    \\ \hline \hline
		\multirow{5}{*}{Arabic}  & Anger   & 0.5    \\ \cline{2-3} 
		& Fear    & 0.0    \\ \cline{2-3} 
		& Sadness & 0.5    \\ \cline{2-3} 
		& Joy     & 0.4    \\ \cline{2-3} 
		& Valence & 0.7    \\ \hline
	\end{tabular}
	\caption{The value of $\alpha$ for each individual model.}
\end{table}

Table 2 shows the value of $\alpha$ for each individual model. All these values were obtained by grid search on the development set.

\subsection{Decision Tree for Ordinal Classification Tasks}
\begin{figure}[h]
	\centering
	\includegraphics[width=0.5\textwidth]{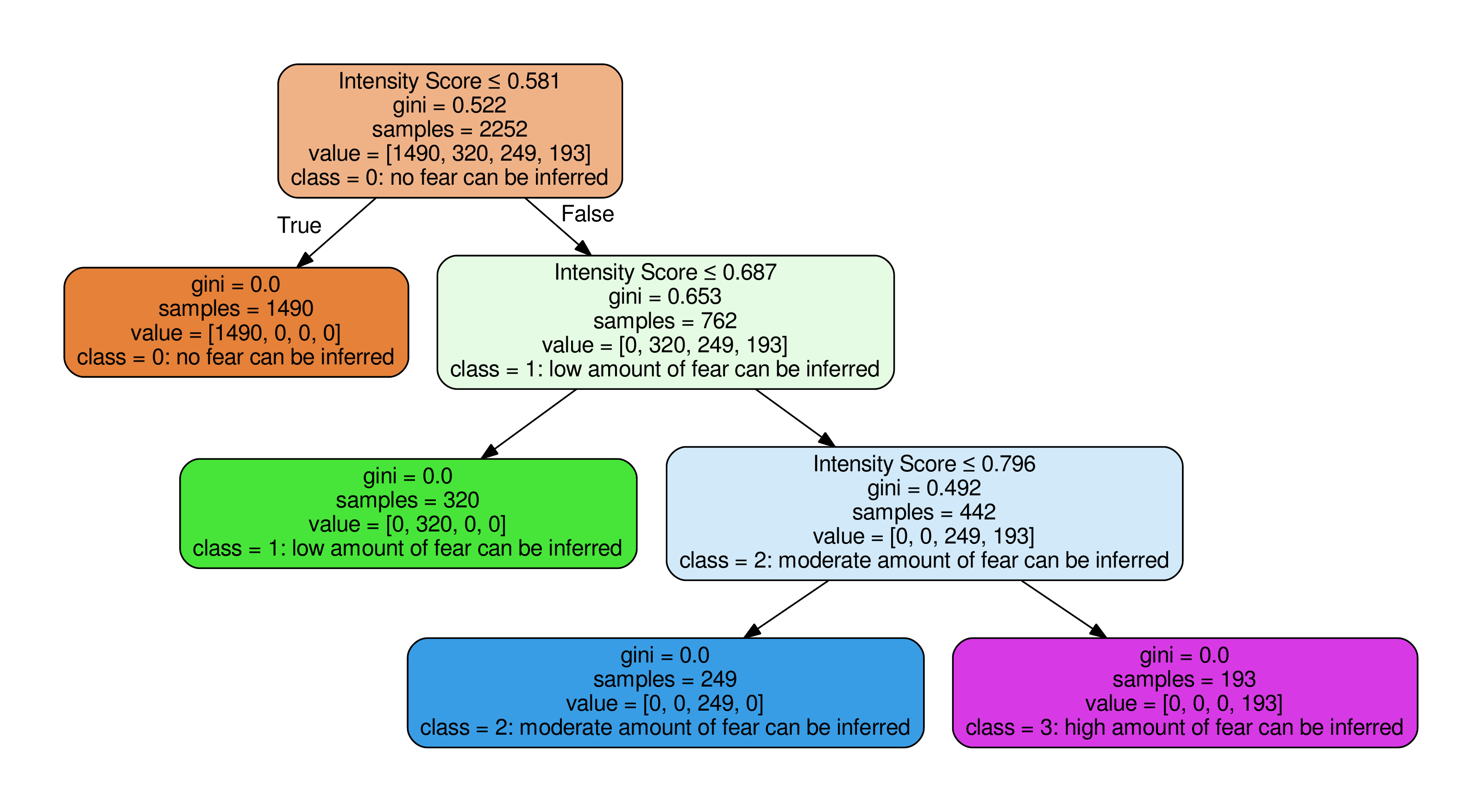}
	\caption{An example of a decision tree classifier.}
	\label{dtfear}
\end{figure}

To solve the problem of ordinal classification we simply used the proposed model as feature extractor and trained a Decision Tree. The idea is to use the emotion/intensity as input feature and use rules generated from the Decision Tree to get the appropriate class. Figure \ref{dtfear} shows as an example the Decision Tree classifier of the $fear$ emotion.

\section{Results}
\label{secer}
We trained and validated our models on the training and validation sets provided by the organizers. More details about the data and the evaluation metrics can be found in \cite{SemEval2018Task1,LREC18-TweetEmo}.

\begin{table*}[h]
	\centering
	
	\label{tblres1}
	\resizebox{2\columnwidth}{!}{%
		\begin{tabular}{|l|l|l|l|l|l|l|l|l|l|l|l|}
			\hline
			\multicolumn{2}{|l|}{\multirow{2}{*}{}}    & \multicolumn{5}{l|}{Pearson (all instances)}  & \multicolumn{5}{l|}{Pearson5 (gold in 0.5-1)}   \\ \cline{3-12} 
			\multicolumn{2}{|l|}{}                     & macro-avg & anger  & fear  & joy    & sadness & macro-avg & anger  & fear   & joy    & sadness \\ \hline
			\multirow{5}{*}{Eng.} & N-Channels ConvNet & 0.712     & 0.713  & 0.725 & 0.718  & 0.692   & 0.538     & 0.575  & 0.502  & 0.519  & 0.555   \\ \cline{2-12} 
			& XGBoost Regressor  & 0.653     & 0.674  & 0.644 & 0.625  & 0.668   & 0.503     & 0.563  & 0.455  & 0.437  & 0.555   \\ \cline{2-12} 
			& Ensemble Model     & 0.724     & 0.731  & 0.733 & 0.722  & 0.711   & 0.560     & 0.606  & 0.522  & 0.525  & 0.587   \\ \cline{2-12} 
			& SVM Unigrams       & 0.520     & 0.526  & 0.525 & 0.575  & 0.453   & 0.396     & 0.455  & 0.302  & 0.476  & 0.350   \\ \cline{2-12} 
			& Random Baseline    & -0.008    & -0.018 & 0.024 & -0.058 & 0.020   & -0.048    & -0.088 & -0.011 & -0.032 & -0.059  \\ \hline
			\multirow{5}{*}{Ara.} & N-Channels ConvNet &    0,655      &  0.639       & 0.628      & 0.705        & 0.648        &     0,516      & 0.473       &   0.605     &0.465        & 0.520        \\ \cline{2-12} 
			& XGBoost Regressor  & 0.596          & 0.494       & 0.540      & 0.713       &  0.637       &           &0.376        &   0.492     &  0.449      & 0.540        \\ \cline{2-12} 
			& Ensemble Model     & 0.667     & 0.627  & 0.627 & 0.738  & 0.675   & 0.533     & 0.479  & 0.604  & 0.490  & 0.560   \\ \cline{2-12} 
			& SVM Unigrams       & 0.455     & 0.406  & 0.435 & 0.530  & 0.450   & 0.353     & 0.344  & 0.366  & 0.332  & 0.367   \\ \cline{2-12} 
			& Random Baseline    & 0.013     & -0.006 & 0.016 & -0.010 & 0.052   & -0.007    & 0.002  & 0.007  & 0.011  & -0.048  \\ \hline
		\end{tabular}
	}
	\caption{EI-reg task results.}
\end{table*}

\begin{table}[h]
	\centering
	
	\label{tblres3}
	\resizebox{\columnwidth}{!}{%
		\begin{tabular}{|l|l|l|l|}
			\hline
			&                    & Pearson & Pearson5 \\ \hline
			\multirow{5}{*}{Eng.} & N-Channels ConvNet &0.825         & 0.645         \\ \cline{2-4} 
			& XGBoost Regressor  & 0.768        & 0.598         \\ \cline{2-4} 
			& Ensemble Model     & 0.828   & 0.658    \\ \cline{2-4} 
			& SVM Unigrams       & 0.585   & 0.449    \\ \cline{2-4} 
			& Random Baseline    & 0.031   & 0.012    \\ \hline
			\multirow{5}{*}{Ara.} & N-Channels ConvNet &   0.817      &   0.550       \\ \cline{2-4} 
			& XGBoost Regressor  &  0.774       &   0.571       \\ \cline{2-4} 
			& Ensemble Model     & 0.828   & 0.578    \\ \cline{2-4} 
			& SVM Unigrams       & 0.571   & 0.423    \\ \cline{2-4} 
			& Random Baseline    & -0.052  & 0.022    \\ \hline
		\end{tabular}
	}
	\caption{V-reg task results.}
\end{table}

Tables 3 and 4 show the results of the emotion and valence intensity regression tasks of our two systems and their combination (the ensemble model). It also shows the baseline results. 
The evaluation metrics are the Pearson correlation for all samples and for a subset of the test set that includes only those tweets with intensity score greater or equal to 0.5. The values in the tables show the superiority of the N-Channels ConvNet over the XGBoost regressor. For instance, the results of the English version of the emotion intensity task show that the N-Channels ConvNet outperforms the XGBoost regressor by 5.9\% with respect to macro-avg measure.  The performance of N-Channels Convnet is very close to the ensemble model. The improvement is only 1.2\%. The improvement in the final system of the Arabic version is very small (0.3\%). The results of the Pearson correlation of samples whose intensity score is greater or equal to 0.5 show that our system can be used as a classifier. This conclusion is confirmed by the results of the ordinal  classification tasks, shown in Tables 5 and 6.

As we described in subsection 3.5, our approach to design a system to solve the ordinal classification tasks was to use the intensity score as input feature to train a Decision Tree. During the inference phase we used our system to produce the intensity score for the new (unseen) samples (i.e. use it as feature extractor). Thus, the performance in this phase heavily relies on the performance of the proposed system. This is clearly shown in the results reported in tables 5 and 6. For example, our system gives very good results in the valence intensity regression task for both the English and Arabic versions (the Pearson correlation is 0.828 for both). This affects positively the performance of our system for the valence ordinal classification tasks (the Pearson correlation is about 0.80 for both).

% Please add the following required packages to your document preamble:
% \usepackage{multirow}
\begin{table*}[h]
	\centering
	
	\label{tblres2}
	\resizebox{2\columnwidth}{!}{%
	\begin{tabular}{|l|l|l|l|l|l|l|l|l|l|l|l|}
		\hline
		\multicolumn{2}{|l|}{\multirow{2}{*}{}}  & \multicolumn{5}{l|}{Pearson}                                       & \multicolumn{5}{l|}{Kappa}                                         \\ \cline{3-12} 
		\multicolumn{2}{|l|}{} & macro-avg   & anger       & fear        & joy        & sadness     & macro-avg   & anger       & fear        & joy        & sadness     \\ \hline
		\multirow{3}{*}{Eng.}  & Our system      & 0.633 (6)   & 0.651 (5)   & 0.595 (2)   & 0.651 (8)  & 0.636 (6)   & 0.608 (3)   & 0.619 (4)   & 0.574 (3)   & 0.607 (10) & 0.632 (4)   \\ \cline{2-12} 
		& SVM Unigrams    & 0.394 (26)  & 0.382 (27)  & 0.355 (26)  & 0.469 (26) & 0.370 (29)  & 0.385 (26)  & 0.375 (26)  & 0.331 (25)  & 0.465 (25) & 0.370 (28)  \\ \cline{2-12} 
		& Random Baseline & -0.016 (37) & -0.062 (38) & 0.047 (33)  & 0.014 (35) & -0.061 (37) & -0.017 (38) & -0.058 (38) & 0.035 (32)  & 0.014 (35) & -0.057 (37) \\ \hline
		\multirow{3}{*}{Ara.}  & Our system      & 0.574 (2)   & 0.572 (1)   & 0.529 (2)   & 0.634 (1)  & 0.563 (3)   & 0.542 (2)   & 0.547 (1)   & 0.516 (2)   & 0.588 (2)  & 0.518 (3)   \\ \cline{2-12} 
		& SVM Unigrams    & 0.542 (2)   & 0.315 (6)   & 0.281 (7)   & 0.281 (6)  & 0.396 (6)   & 0.299 (6)   & 0.276 (7)   & 0.249 (6)   & 0.385 (6)  & 0.287 (7)   \\ \cline{2-12} 
		& Random Baseline & 0.006 (12)  & -0.057 (12) & -0.019 (12) & 0.008 (12) & 0.092 (11)  & 0.006 (12)  & -0.057 (14) & -0.019 (12) & 0.007 (12) & 0.091 (10)  \\ \hline
	\end{tabular}
}
\caption{EI-oc task results.}
\end{table*}

% Please add the following required packages to your document preamble:
% \usepackage{multirow}
\begin{table}[h]
	\centering
	
	\label{tblres4}
	\begin{tabular}{|l|l|l|l|}
		\hline
		&                 & Pearson & Kappa  \\ \hline
		\multirow{3}{*}{Eng.} & Our system      & 0.796   & 0.791  \\ \cline{2-4} 
		& SVM Unigrams    & 0.509   & 0.504  \\ \cline{2-4} 
		& Random Baseline & -0.010  & -0.010 \\ \hline
		\multirow{3}{*}{Ara.} & Ensemble Model  & 0.809   & 0.783  \\ \cline{2-4} 
		& SVM Unigrams    & 0.471   & 0.470  \\ \cline{2-4} 
		& Random Baseline & 0.011   & 0.011  \\ \hline
	\end{tabular}
	\caption{V-oc task results.}
\end{table}

%The evaluation metrics used by the task organizers were the macroaveraged recall ($\rho$), the F1 averaged across the positives and the negatives $F1^{PN}$ and the accuracy ($Acc$) \cite{SemEval:2017:task4}.

%The system has been tested on 12,284 English-language tweets and 6100 Arabic-language tweets provided by the organizers. The golden answers of all the test tweets were omitted by the organizers. The official evaluation results of our system are reported along with the top 10 systems and the baseline results in Table 2 and 3. Our system ranks 8th among 38 systems in the English-language tweets and ranks 2nd among 8 systems in the Arabic language tweets. The baselines 1, 2 and 3 stand for case when the system classify all the tweets as positive, negative and neutral respectively.

\section{Conclusion}
\label{secc}
We have presented an ensemble model of  two different approaches. The first one, called N-Channels ConvNet, is a deep learning approach whereas the second one is an XGBoost regressor based on a set of embedding and lexicons-based features. The ensemble technique helped to improve the performance of the final model in all subtasks. We have realized that The N-Channels ConvNet gives a performance very close to the ensemble model. This observation confirms the fact that deep learning models, and especially ConvNets, have achieved remarkable results in many fields such as computer vision, speech recognition and natural language processing. Distant Supervision is an approach of transfer learning which aims to train a model on a large amount of semi-labeled data and use it as a pre-trained model for training another model on a small amount of fully-labeled data. This approach has been shown to be very efficient. Thus, the authors are considering the possibility of using this technique to improve the proposed system.   

\section*{Acknowledgment}
This work was partially supported by URV Research Support Funds (2017PFR-URV-B2-61 and  Mart\'{\i} i Franqu\'es PhD grant).

\bibliography{semeval2017}

\begin{thebibliography}{24}
\expandafter\ifx\csname natexlab\endcsname\relax\def\natexlab#1{#1}\fi

\bibitem[{Baziotis et~al.(2017)Baziotis, Pelekis, and Doulkeridis}]{baziotis17}
Christos Baziotis, Nikos Pelekis, and Christos Doulkeridis. 2017.
\newblock Datastories at semeval-2017 task 4: Deep lstm with attention for
  message-level and topic-based sentiment analysis.
\newblock In \emph{Proceedings of the 11th International Workshop on Semantic
  Evaluation (SemEval-2017)}, pages 747--754, Vancouver, Canada. Association
  for Computational Linguistics.

\bibitem[{Chen and Guestrin(2016)}]{xgb}
Tianqi Chen and Carlos Guestrin. 2016.
\newblock \href {http://arxiv.org/abs/1603.02754} {{XGBoost: {A} Scalable Tree
  Boosting System}}.
\newblock \emph{CoRR}, abs/1603.02754.

\bibitem[{Choi and Wiebe(2014)}]{choi2014+}
Yoonjung Choi and Janyce Wiebe. 2014.
\newblock +/-effectwordnet: Sense-level lexicon acquisition for opinion
  inference.
\newblock In \emph{Proceedings of the 2014 Conference on Empirical Methods in
  Natural Language Processing (EMNLP)}, pages 1181--1191.

\bibitem[{Gimpel et~al.(2011)Gimpel, Schneider, O'Connor, Das, Mills,
  Eisenstein, Heilman, Yogatama, Flanigan, and Smith}]{Gimpel2011}
Kevin Gimpel, Nathan Schneider, Brendan O'Connor, Dipanjan Das, Daniel Mills,
  Jacob Eisenstein, Michael Heilman, Dani Yogatama, Jeffrey Flanigan, and
  Noah~A. Smith. 2011.
\newblock {Part-of-speech Tagging for Twitter: Annotation, Features, and
  Experiments}.
\newblock In \emph{Proceedings of the 49th Annual Meeting of the Association
  for Computational Linguistics: Human Language Technologies: Short Papers -
  Volume 2}, HLT '11, pages 42--47, Stroudsburg, PA, USA. Association for
  Computational Linguistics.

\bibitem[{Green and Manning(2010)}]{green2010better}
Spence Green and Christopher~D Manning. 2010.
\newblock Better arabic parsing: Baselines, evaluations, and analysis.
\newblock In \emph{Proceedings of the 23rd International Conference on
  Computational Linguistics}, pages 394--402. Association for Computational
  Linguistics.

\bibitem[{Hu and Liu(2004)}]{hu2004mining}
Minqing Hu and Bing Liu. 2004.
\newblock {Mining and summarizing customer reviews}.
\newblock In \emph{Proceedings of the tenth ACM SIGKDD international conference
  on Knowledge discovery and data mining}, pages 168--177. ACM.

\bibitem[{Jabreel and Moreno(2016)}]{JabreelM16}
Mohammed Jabreel and Antonio Moreno. 2016.
\newblock Sentirich: Sentiment analysis of tweets based on a rich set of
  features.
\newblock In \emph{Artificial Intelligence Research and Development -
  Proceedings of the 19th International Conference of the Catalan Association
  for Artificial Intelligence, Barcelona, Catalonia, Spain, October 19-21,
  2016}, pages 137--146.

\bibitem[{Kim(2014)}]{kim2014convolutional}
Yoon Kim. 2014.
\newblock Convolutional neural networks for sentence classification.
\newblock \emph{arXiv preprint arXiv:1408.5882}.

\bibitem[{Mikolov et~al.(2013)Mikolov, Chen, Corrado, and
  Dean}]{mikolov2013efficient}
Tomas Mikolov, Kai Chen, Greg Corrado, and Jeffrey Dean. 2013.
\newblock Efficient estimation of word representations in vector space.
\newblock \emph{arXiv preprint arXiv:1301.3781}.

\bibitem[{Mohammad et~al.(2013)Mohammad, Kiritchenko, and Zhu}]{MohammadKZ2013}
Saif Mohammad, Svetlana Kiritchenko, and Xiaodan Zhu. 2013.
\newblock {NRC-Canada: Building the State-of-the-Art in Sentiment Analysis of
  Tweets}.
\newblock In \emph{Proceedings of the seventh international workshop on
  Semantic Evaluation Exercises (SemEval-2013)}, Atlanta, Georgia, USA.

\bibitem[{Mohammad et~al.(2016)Mohammad, Salameh, and
  Kiritchenko}]{arabiclex2016}
Saif Mohammad, Mohammad Salameh, and Svetlana Kiritchenko. 2016.
\newblock Sentiment lexicons for arabic social media.
\newblock In \emph{Proceedings of 10th edition of the the Language Resources
  and Evaluation Conference (LREC)}, Portoro\v{z}, Slovenia.

\bibitem[{Mohammad(2016)}]{SentimentEmotionSurvey2015}
Saif~M. Mohammad. 2016.
\newblock Sentiment analysis: Detecting valence, emotions, and other affectual
  states from text.
\newblock In Herb Meiselman, editor, \emph{Emotion Measurement}. Elsevier.

\bibitem[{Mohammad and Bravo{-}M\'arquez(2017)}]{wassa17}
Saif~M. Mohammad and Felipe Bravo{-}M\'arquez. 2017.
\newblock \href {http://arxiv.org/abs/1708.03700} {{WASSA-2017} shared task on
  emotion intensity}.
\newblock \emph{CoRR}, abs/1708.03700.

\bibitem[{Mohammad et~al.(2018)Mohammad, Bravo-Marquez, Salameh, and
  Kiritchenko}]{SemEval2018Task1}
Saif~M. Mohammad, Felipe Bravo-Marquez, Mohammad Salameh, and Svetlana
  Kiritchenko. 2018.
\newblock Semeval-2018 {T}ask 1: {A}ffect in tweets.
\newblock In \emph{Proceedings of International Workshop on Semantic Evaluation
  (SemEval-2018)}, New Orleans, LA, USA.

\bibitem[{Mohammad and Kiritchenko(2018)}]{LREC18-TweetEmo}
Saif~M. Mohammad and Svetlana Kiritchenko. 2018.
\newblock Understanding emotions: A dataset of tweets to study interactions
  between affect categories.
\newblock In \emph{Proceedings of the 11th Edition of the Language Resources
  and Evaluation Conference}, Miyazaki, Japan.

\bibitem[{Nielsen(2011)}]{afinn}
Finn~Årup Nielsen. 2011.
\newblock \href
  {http://dblp.uni-trier.de/db/journals/corr/corr1103.html#abs-1103-2903} {A
  new anew: Evaluation of a word list for sentiment analysis in microblogs}.
\newblock \emph{CoRR}, abs/1103.2903.

\bibitem[{Rouvier and Favre(2016)}]{rouvier2016sensei}
Mickael Rouvier and Benoit Favre. 2016.
\newblock Sensei-lif at semeval-2016 task 4: Polarity embedding fusion for
  robust sentiment analysis.
\newblock In \emph{Proceedings of the 10th International Workshop on Semantic
  Evaluation (SemEval 2016), San Diego, US}.

\bibitem[{Severyn and Moschitti(2015)}]{severyn2015unitn}
Aliaksei Severyn and Alessandro Moschitti. 2015.
\newblock {UNITN: Training deep convolutional neural network for Twitter
  sentiment classification}.
\newblock In \emph{Proceedings of the 9th International Workshop on Semantic
  Evaluation (SemEval 2015), Association for Computational Linguistics, Denver,
  Colorado}, pages 464--469.

\bibitem[{Soliman et~al.(2017)Soliman, Eissa, and El-Beltagy}]{SOLIMAN2017256}
Abu~Bakr Soliman, Kareem Eissa, and Samhaa~R. El-Beltagy. 2017.
\newblock \href {https://doi.org/https://doi.org/10.1016/j.procs.2017.10.117}
  {Aravec: A set of arabic word embedding models for use in arabic nlp}.
\newblock \emph{Procedia Computer Science}, 117:256 -- 265.
\newblock Arabic Computational Linguistics.

\bibitem[{Stone et~al.(1968)Stone, Dunphy, Smith, and
  Ogilvie}]{stone1968general}
Philip Stone, Dexter~C Dunphy, Marshall~S Smith, and DM~Ogilvie. 1968.
\newblock {The general inquirer: A computer approach to content analysis}.
\newblock \emph{Journal of Regional Science}, 8(1):113--116.

\bibitem[{Szegedy et~al.(2016)Szegedy, Ioffe, and Vanhoucke}]{inception4}
Christian Szegedy, Sergey Ioffe, and Vincent Vanhoucke. 2016.
\newblock \href {http://arxiv.org/abs/1602.07261} {{Inception-v4,
  Inception-ResNet and the Impact of Residual Connections on Learning}}.
\newblock \emph{CoRR}, abs/1602.07261.

\bibitem[{Tang et~al.(2014{\natexlab{a}})Tang, Wei, Qin, Liu, and
  Zhou}]{tang2014Coooolll}
Duyu Tang, Furu Wei, Bing Qin, Ting Liu, and Ming Zhou. 2014{\natexlab{a}}.
\newblock \href {http://www.aclweb.org/anthology/S14-2033} {{Coooolll: A Deep
  Learning System for Twitter Sentiment Classification}}.
\newblock In \emph{Proceedings of the 8th International Workshop on Semantic
  Evaluation (SemEval 2014)}, pages 208--212, Dublin, Ireland. Association for
  Computational Linguistics and Dublin City University.

\bibitem[{Tang et~al.(2014{\natexlab{b}})Tang, Wei, Yang, Zhou, Liu, and
  Qin}]{tang2014SSWE}
Duyu Tang, Furu Wei, Nan Yang, Ming Zhou, Ting Liu, and Bing Qin.
  2014{\natexlab{b}}.
\newblock \href {http://www.aclweb.org/anthology/P14-1146} {{Learning
  Sentiment-Specific Word Embedding for Twitter Sentiment Classification}}.
\newblock In \emph{Proceedings of the 52nd Annual Meeting of the Association
  for Computational Linguistics (Volume 1: Long Papers)}, pages 1555--1565,
  Baltimore, Maryland. Association for Computational Linguistics.

\bibitem[{Zahran et~al.(2015)Zahran, Magooda, Mahgoub, Raafat, Rashwan, and
  Atyia}]{zahran2015word}
Mohamed~A Zahran, Ahmed Magooda, Ashraf~Y Mahgoub, Hazem Raafat, Mohsen
  Rashwan, and Amir Atyia. 2015.
\newblock Word representations in vector space and their applications for
  arabic.
\newblock In \emph{International Conference on Intelligent Text Processing and
  Computational Linguistics}, pages 430--443. Springer.

\end{thebibliography}
\bibliographystyle{acl_natbib}

\end{document}